# The Biometric Menagerie – A Fuzzy and Inconsistent Concept

N. Popescu-Bodorin[1], V.E. Balas[2], I.M. Motoc[1]

*Abstract—* This paper proves that in iris recognition, the concepts of *sheep, goats, lambs* and *wolves* - as proposed by Doddington and Yager in the so-called Biometric Menagerie, are at most fuzzy and at least not quite well defined. They depend not only on the users or on their biometric templates, but also on the parameters that calibrate the iris recognition system. This paper shows that, in the case of iris recognition, the extensions of these concepts have very unsharp and unstable (non-stationary) boundaries. The membership of a user to these categories is more often expressed as a degree (as a fuzzy value) rather than as a crisp value. Moreover, they are defined by fuzzy Sugeno rules instead of classical (crisp) definitions. For these reasons, we said that the Biometric Menagerie proposed by Doddington and Yager could be at most a fuzzy concept of biometry, but even this status is conditioned by improving its definition. All of these facts are confirmed experimentally in a series of 12 exhaustive iris recognition tests undertaken for University of Bath Iris Image Database while using three different iris code dimensions (256x16, 128x8 and 64x4), two different iris texture encoders (Log-Gabor and Haar-Hilbert) and two different types of safety models.

## I. Introduction

WHILE working around speech recognition, Doddington et al. introduced in [2] four concepts reflecting four types of users: *sheep, goats, lambs* and *wolves* – which together form the so-called Biometric Menagerie. The second section of this paper presents an objective critique of this concept.

As far as we know, in 2010, N. Yager et al. [12] generalized Doddington's classification (also known as Doddington's zoo) for all fields of biometrics. Since then, just two papers investigating the presence of sheep, goats, lambs and wolves in certain benchmark databases have been published.

After [7] and [4], this is the third paper that analyses the partitioning of the iris code space extracted for a certain database (University of Bath Iris Image Database, UBIID, [10] – in our case) as a Fuzzy Biometric Menagerie showing that the extensions of the concepts *wolf, lambs, sheep* and *goats* have very unsharp and unstable (non stationary) boundaries. Moreover, the membership of a user to these categories can be more often expressed as a degree (as a fuzzy value) rather than as a crisp value. The fact that the Biometric Menagerie could be a fuzzy concept is confirmed experimentally here in a series of 12 exhaustive iris recognition tests undertaken for UBIID [10] by using three different iris code dimensions (256x16, 128x8 and 64x4), two different iris texture encoders (Log-Gabor and Haar-Hilbert [6]) and two different types of safety models [7]. All of these tests illustrate that the partitioning of template-space accordingly to the fuzzy concepts *wolves, lambs, sheep*, and *goats* depends not only on the users or on their biometric templates, but also on the parameters that calibrate the iris recognition system – fact which is also confirmed in [3] for a different iris image database (Iris Challenge Evaluation, [3]).

## II. 'Biometric Menagerie' in iris recognition. Open problems and contradictory issues.

Doddington et al. [2] and Yager et al. [12] defined the concepts of *sheep-user, goat-user, lamb-use*r and *wolf-user* as follows:

*Definition 1* (Yager, [12]):
- The *sheep* are those users for which the similarity score is high for genuine comparisons and low for imposter comparisons;
- The *goats* are those users which, most of the time, obtain low similarity scores for genuine comparisons;
- The *lambs* are those users easy to imitate (by *wolves*) and for which the similarity score for imposter comparison can be relatively high.
- The *wolves* are those users particularly good at impersonating other users (or in other words, as Yager said, the *wolves* "prey upon *lambs*" [12]) obtaining relatively high similarity scores for imposter comparison between them and the lambs.

### A. Classifying users vs classifying templates

Firstly, anyone should remark (we certainly did it) that classifying users in the first place is not necessarily a very good idea, simply because, any claimed relation that possibly hold two users or more is caused by something that happens with certain binary biometric templates stored in the system on their name. What happens with the templates determines what

[1] Nicolaie Popescu-Bodorin, Iulia M. Motoc, IEEE Members, Artificial Intelligence & Computational Logic Lab., Mathematics & Computer Science Department, Spiru Haret University, Bucharest Romania; Emails: bodorin, motoc [at] irisbiometrics.org
[2] Valentina E. Balas, IEEE Senior Member, Associate Professor, Automatics and Applied Software Department, Faculty of Engineering, Aurel Vlaicu University, Arad, Romania, Email: balas [at] drbalas.ro

happens with the users, not vice versa. Hence, in any biometric system (including those based on iris recognition), the natural approach to classifying users goes through classifying biometric templates (through classifying iris codes - in our particular case). Therefore, a correct foundation for a hypothetically objective model called Biometric Menagerie should start with defining the '*animals*' [12] by analyzing their hypostases, i.e. in terms of biometric templates:

*Definition 2*:
- The *sheep-templates* are those for which the similarity scores associated to their genuine comparisons are *high enough* and the similarity scores associated to their imposter comparisons are *low enough* such that a safety threshold or a safety interval to separate the two distributions of genuine and imposter scores computed for them;
- The *goat-templates* are those that, *most of the time* or *too often*, obtain low similarity scores for their genuine comparisons;
- The *lamb-templates* are those *easy to imitate* (by wolves) and for which the similarity scores associated to their imposter comparisons can be relatively high;
- The *wolf-templates* are those particularly good at matching lamb-templates, obtaining *relatively high* similarity scores for imposter comparison between them and their pray (lamb-templates);
- Biometric Menagerie is a partitioning of biometric template space into the four classes defined above.

*B. Fuzzy Biometric Menagerie vs System Calibration*

Secondly, even admitting the fact that Biometric Menagerie is a well-defined concept, all conditions expressed in the above two definitions are rather fuzzy if-then Sugeno rules [11] than regular conditions of a classical definition – i.e. conditions on *genus and differentia* that do not contain fuzzy elements. More precisely, both definitions are intensional, the genus being the space of biometric templates, whereas a fuzzy rule declares the differentia. Therefore, there is no doubt that Biometric Menagerie is a fuzzy partitioning of the biometric templates space in sub-classes defined as extensions of the fuzzy concepts (pre-images of the fuzzy labels) sheep, goats, lambs and wolves, regardless the fact that it could refer to users or to biometric templates. As an example, let us formalize one condition of the second definition as a fuzzy if-then Sugeno rule:

IF: T is a biometric template associated to *high* genuine scores and *low* imposter scores THEN: T is a sheep-template whose structure is similar to that of a linguistic control rule [11] describing a multi-input & single-output system:

IF: X is *f-label-1* and Y is *f-label-2* THEN: Z is *f-label-3*.

As seen above, the concept of sheep-template is fuzzy and so it is the entire Biometric Menagerie. Despite the fact that the genus of sheep-template is a crisp set, is the fuzzy rule from above that declares the differentia using the fuzzy linguistic labels '*high*' and '*low*' whose possible quantitative semantics correspond to a choice of some underlying fuzzy sets associated with some membership functions. Someone must choose a numerical interpretation of what it means to be *high* as a genuine score and *low* as an imposter score, operation usually referred to as a part of calibrating the biometric system. Therefore, our first hunch (now partially validated through experimental work) was that the Biometric Menagerie is rather depending on the calibration of biometric system than being an objective concept, well defined and applicable in general for the users that pass through different single-biometric systems that use the same biometric trait (iris, face, fingerprint, palm-vein, etc.).

*C. From partitioning templates to partitioning users*

Let us assume that in an iris recognition system we need to define a partitioning of the users according to what happens with their biometric templates. For example, we could consider the case in which a user $U_1$ posses a template $T_1$ that candidates for the role of being a wolf-template by obtaining six imposter similarity scores high enough to generate six false accepts with six different users. In the same system, a user U2 posses the templates $T_2^1$, $T_2^2$, $T_2^3$, each of them obtaining two imposter similarity scores high enough such that together they generate the same number of six false accepts with six different users. As seen in our example, detecting a wolf-user could be a problem of finding a group of template-wolves that together satisfy some conditions. The question is which one of those two users is a wolf-user. The answer hardly depends on a convention that the system use for qualifying users as wolves based on what happens with their templates (taken individually or as a group). At least because it relies on the detection of some wolf-templates - detection done by following a fuzzy rule (as described above), such a convention is a fuzzy if-then rule also:

IF: for the user U there is a group G of its templates satisfying a *well chosen* f-convention *FC* THEN: U is a *wolf-user*

Hence, in the rule described above, besides the fact that the detection of the individual wolf-templates is fuzzy, there are two additional degrees of freedom for interpreting the fuzzy labels "*well chosen*" and "*FC*". This fact makes the process of identifying the wolf-users even fuzzier and more subjective than the process of finding wolf-templates. Consequently, the concept of Biometric Menagerie as introduced by Doddington et al. in [2] and Yager et al. in [12] and even the concept of Biometric Menagerie discussed here in definition 2 are all fuzzy and subjective concepts, regardless if they consist in partitioning users or templates.

The fact itself that the process of partitioning the users or the templates in a Biometric Menagerie is a fuzzy one cannot be negatively connotated by default, excepting, of course, the cases in which there is not enough cointension between this artificial partitioning and the natural tendency of grouping that users actually have in reality. Unfortunately, this is exactly the case here, as shown below.

Biometric recognition is a diachronic process and therefore the basic vocabulary of any recognition theory should refer user instances, i.e. pairs (U, t) where U is a user and t is a time.

A recognition theory is logically consistent if and only if, regardless the time values $t_1$ and $t_2$, the similarity $(U_1, t_1) \equiv (U_2, t_2)$ certainly take place only for the same user $U_1 = U_2$. In other words, all users enrolled in the system diachronically generate a set of genuine comparisons that posses the pattern $(U, t_1)$-to-$(U, t_2)$ and a set of imposter comparisons that also share a common pattern $(U_1, t)$-to-$(U_2, \tau)$ with $U_1 \neq U_2$ (the relation between t and $\tau$ having no importance in this case). Hence, the natural tendency of grouping that user instances actually have points out to only two classes, not to four classes – as the Biometric Menagerie has.

The situation described above is an important example illustrating that *fuzzy* could sometimes mean *logically inconsistent*, such is the case of artificial partitioning of the users in a Biometric Menagerie with four fuzzy classes, while the natural tendency of grouping that the users actually have in a consistent theory of recognition point out to a binary classification.

### D. FBM vs. iris codes space homogeneity

According to the above definitions, the wolves are those users (proved or suspected – depending on how accurate the wolf definition actually is) responsible for much of the False Accept Rate (FAR), whereas the goats are the users responsible for much of the False Reject Rate (FRR). This is why the current paper gives a special attention to these two categories of users.

However, right from this moment it is very clear that accepting the above definitions would mean to accept that some users would be somehow special (more special than others) and therefore, some elements of the iris code space would be somehow more special than others, hence, the question if the iris code space is homogeneous or heterogeneous would certainly appear.

A thing to know for sure is if the iris code space actually is homogeneous or not. We believe it is. The situation described above is a classical kind of example illustrating that when adding something that initially appears inoffensive to a model (like a classification of users – in the current case) actually blows up the foundations of the model by introducing the contradiction in its logic. Let us assume that the iris code space is heterogeneous (i.e. it supports the definition 2) and that the partitioning of iris codes space is cointensive with a corresponding partitioning of user space, which consequently is heterogeneous on its turn. Can anybody tell us what makes the user space heterogeneous in the first place?

In a lottery, many players can win the minor prizes by partially matching the official extracted variant. Hence, we could say that the extracted variant is a wolf hunting on lambs (the winners of the minor prizes). We could say, but we do not say that. Nothing aggregates the group of these winners together, except the pure chance. In the same manner, the odds produce the matching between one specific iris code and many others purely by chance, meaning that the iris code space is locally too agglomerated and this agglomeration could become homogeneously present in the iris code space. The solution is not to invent wolves and lambs, but to recalibrate the system by increasing the power of discrimination between the future biometric templates.

### E. FBM vs. Similarity Score Symmetry

The fact that Biometric Menagerie is fuzzy (regardless it refers to users or templates) is not the worst thing in the world. The real problem is that it is not objective. In order to prove that, let us comment the wolf-lamb relation.

According to Yager et al. [12], wolf-lambs relation is one-to-many, one wolf taking many lambs. However, in a biometric system in which the relation between users (between templates) is symmetric (why should not be?), if the user $U_1$ (the template $T_1$) impersonates the user $U_2$ (the template $T_2$), it is equally true that the user $U_2$ (the template $T_2$) impersonates the user $U_1$ (the template $T_1$), also. Therefore, it is not clear at all who is the hunter and who is hunted. Someone has chosen to say that, most probably (according to some experiences), the wolves take many lambs. Our question is: what if, actually, *many wolves target the same lamb*.

The situation described above allows us to say that denoting some users (templates) as wolves and others as lambs is a pure subjective convention which really affects the objectivity of Biometric Menagerie as a concept.

## III. EXPERIMENTAL RESULTS

This section presents the results of 12 exhaustive iris recognition tests, undertaken on the database [10], using iris codes of dimensions 256x16, 128x8 and 64x4.

All tests use the second version of Circular Fuzzy Iris Segmentation procedure (CFIS2, proposed in [5], available for download in [7]), the iris segments being further normalized to the appropriate dimension and encoded as binary iris codes by using Haar-Hilbert [6] and Log-Gabor [6] texture encoders. Each comparison between iris codes results in a matching score computed as Hamming similarity (unitary complement of Hamming distance). For each test, all-to-all comparisons result in similarity scores further interpreted as being *low* or *high* enough to motivate a biometric decision accordingly to the following two fuzzy if-then Sugeno [11] rules:

| IF: | MS(C) is *low* | THEN: | C is (an) *imposter comparison* |
|---|---|---|---|
| IF: | MS(C) is *high* | THEN: | C is (a) *genuine comparison* |

where MS is the matching score and C is a comparison.

## A. Two paradigms of test scenarios

For each test, the precisiation of the security model assumes the deffuzification of the fuzzy labels '*low*' and '*high*' as intervals situated on the left and right sides relative to a threshold value identified as the abscise of the EER point:

$$t_{EER} = (FAR^{-1}(EER) = FRR^{-1}(EER)),$$

or either relative to a safety interval initialized and determined maximally by the minimum Genuine Score (mGS) and the Maximum Imposter Score (MIS), and further decreased iteratively until the extensions of the f-concepts '*wolf*' and '*lamb*' become populated with some examples of wolf- and lamb-templates, respectively. For a given calibration of the recognition system established in terms of segmentation, normalization and encoding procedures, the safety model corresponding to the second case described above (that using a safety interval) is described by the following fuzzy 3-valent disambiguated model:

| IF: | MS(C) is *under* the safety band | THEN: | C is an *imposter comparison* |
|---|---|---|---|
| IF: | MS(C) is *within* the safety band | THEN: | C is *undecidable* |
| IF: | MS(C) is *above* the safety band | THEN: | C is a *genuine comparison* |

## B. The dynamics of FBM. The first and the last wolves and goats

If the safety band is maximal - i.e. the safety band is the interval [mGS, MIS], all the comparisons within $MS^{-1}$ ([mGS, MIS]) are undecidable and therefore there are no wolfs, no lambs and no goats in the system, all users and templates qualifying as sheep. When the safety band narrows from both sides toward the threshold corresponding to the experimentally determined EER point, the examples of wolf-, lamb- and goat-templates slightly came into view. For this reason, we called these kind of templates *marginal wolf-, lamb- and goat-templates*. They are the first wolves, lambs and goats that appear in the system when the level of security decreases from the maximal safety band toward the threshold $t_{EER}$. The idea of searching for wolves and goats while the safety band narrows toward $t_{EER}$ allow us to analyze the dynamics of Biometric Menagerie along the process of decreasing the safety level in a balanced manner that negotiates between false accepts and false rejects. Besides, in order to compare the partitioning of the users/templates in two different iris recognition systems, it was necessary to identify functioning regimes in which the two systems are objectively comparable. We found two functioning regimes of this kind: one identified through the maximal safety band [mGS, MIS] and other identified through $t_{EER}$. These two functioning regimes are the extreme cases between which anyone can study the variability of Biometric Menagerie while the safety band converges to $t_{EER}$ through hypostases that balance the FAR-FRR risks. Safety band hypostases together simulate a family of decreasing nested Cantor intervals allowing us to see the stabilization of the Biometric Menagerie as a process of convergence, along which different iris recognition system are comparable. The last interval of this family is the smallest (first) in the order of inclusion and the last in the order given by the balanced risks assumed in the system. For this reason, we called the members of Biometric Menagerie detected when the system runs at EER, as being the last ones (*last wolf-, lamb- and goat-templates*). They are the last detected of their kind when system security falls in a balanced manner to the EER. All of these things allow us to state the following definition:

*Definition 3*: Let us consider an iris recognition system in which the score distributions overlap each other. Then:
- the *first wolf-, lamb- and goat-templates* are those detected when the system is running at the security level given by the first fuzzy 3-valent disambiguated model [7] in which they appear when the maximal safety band [mGS, MIS] narrows to $t_{EER}$ such that to keep FAR-FRR risks balanced.
- the *last wolf-, lamb- and goat-templates* are those detected when the system is running at EER (i.e. the system is running on that safety threshold which balances the FAR-FRR risks).

## C. Two series of tests

The first series of six tests aims to identify the indices of the *first wolf-* and *goat-templates* detected when running the system with different encoders (Haar-Hilbert and Log-Gabor), with different iris code dimensions (256x16, 128x8, 64x4), at a high security level given by that safety band who allows the wolves and the goats to appear in the system. Table 1 shows the values determining the safety bands detected for each of these tests.

TABLE I
.THE SAFETY BANDS AND THEIR WIDTH FOR THE FIRST SERIES OF SIX ALL-TO-ALL IRIS RECOGNITION TESTS

| Iris code dimension | | 64x4 | 128x8 | 256x16 |
|---|---|---|---|---|
| Log-Gabor encoder | Safety band | [0.6003, 0.9075] | [0.6277, 0.6555] | [0.5566, 0.5757] |
| | Width | 0.3072 | 0.0278 | 0.0191 |
| Haar-Hilbert encoder | Safety Band | [0.6091, 0.6722] | [0.5456, 0.6823] | [0.5224, 0.5467] |
| | Width | 0.0631 | 0.1367 | 0.0243 |

The second series of six tests has the same purposes as the first one, but each time the system is running at a maximally acceptable balanced degradation of the security level given by functioning at EER threshold ($t_{EER}$). Table 2 shows the values determining the safety bands detected for each of these tests.

TABLE II
. THE EER AND $t_{EER}$ FOR THE SECOND SERIES OF SIX ALL-TO-ALL IRIS RECOGNITION TESTS

| Iris code dimension | | 64x4 | 128x8 | 256x16 |
|---|---|---|---|---|
| Log-Gabor encoder | EER | 4.08E-2 | 9.37E-4 | 6.03E-4 |
| | $t_{EER}$ | 0.7529 | 0.6392 | 0.5686 |
| Haar-Hilbert encoder | EER | 8.60E-3 | 1.70E-3 | 2.30E-3 |
| | $t_{EER}$ | 0.6471 | 0.5765 | 0.5490 |

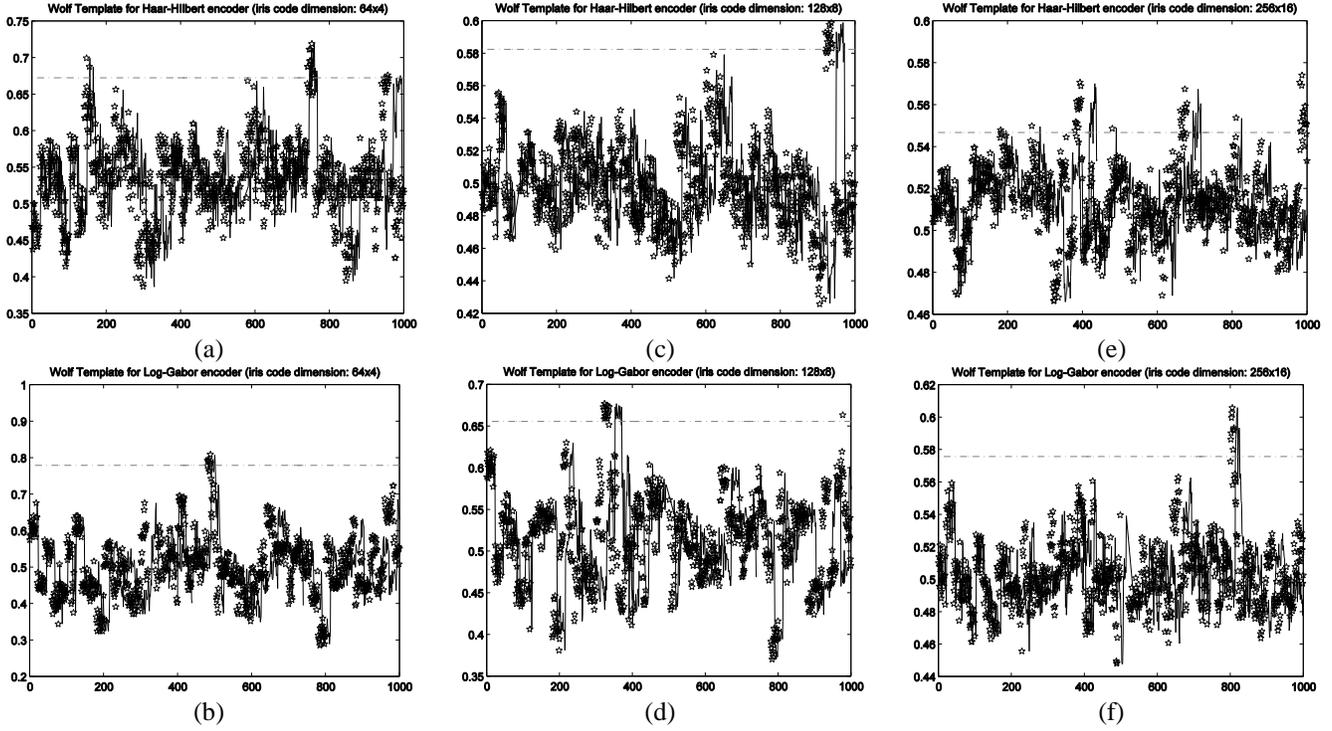

Fig. 1. The *marginal wolf-templates* obtained for Haar-Hilbert (64x4 – a, 128x8 – c, 256x16 – e) and Log-Gabor (64x4 – b, 128x8 – d, 256x16 – f) encoders.

As seen in Table 2, accordingly to the EER criterion, the best calibration of the iris recognition system is that one using iris segments of dimension 256x16 and based on Log-Gabor encoder (EER = 6.0265E-4).

Also, the best calibration presented in Table 1 is that one having the smallest overlapping between the two score distributions, namely that one using iris segments of dimension 256x16 and based on Log-Gabor encoder (for which the amplitude of the overlapping is 0.0191).

### D. Detecting the marginal wolf and goat templates

We recall that the safety bands used in the first series of six iris recognition tests are adaptively determined by narrowing the maximal safety band [mGS, MIS] toward $t_{EER}$ while keeping the FAR-FRR risks balanced, until some examples of wolf and goat templates appear in the system (ensuring that the extensions of the corresponding concepts are not empty). Hence, each test results in a set containing the *first* (the *marginal*) *goat-* and *wolf-templates* corresponding to a given calibration of the biometric system in terms of encoder and iris code size.

Fig. 1 illustrates the fact that although the iris code dimension increases, the number of impersonations oscillates when using Log-Gabor encoder, and increases when using Haar-Hilbert encoder. As seen by comparing Fig. 1.a and Fig. 1.b (both of them obtained for the iris codes of dimension 64x4), the number of cases of impersonation was higher for the wolf-template obtained for Haar-Hilbert encoder than the one obtained for Log-Gabor encoder.

For iris codes of dimension 128x8 (Fig. 1.c and Fig. 1.d), the number of impersonations obtained when using Haar-Hilbert encoder is smaller than when using Log-Gabor encoder. For iris code of dimension 256x16, the Haar-Hilbert encoder obtained the greatest number of impersonations, as we can observe also by comparing the behavior of the wolf templates represented in Fig. 1.e and Fig. 1.f.

TABLE III
THE *MARGINAL WOLF-/GOAT-TEMPLATES* OBTAINED BY FINDING THE CORRESPONDING SAFETY BAND

| Iris code dimension<br>Template type | | 64x4<br>Wolf \| Goat | 128x8<br>Wolf \| Goat | 256x16<br>Wolf \| Goat |
|---|---|---|---|---|
| Log-Gabor encoder | Number of comparisons | 7 \| 4 | 17 \| 3 | 9 \| 3 |
| | Template's index | **334 \| 496** | **484 \| 475** | **505 \| 565** |
| Haar-Hilbert encoder | Number of comparisons | 15 \| 3 | 15 \| 3 | 46 \| 4 |
| | Template's index | **549 \| 565** | **88 \| 565** | **236 \| 565** |

Table 3 presents the results obtained in these six tests performed to find the *marginal wolf-templates*. As seen in Table 3, each test points out to a different *marginal wolf-template* (which is an experimental result that agrees to those presented in [4] for the wolves detected in ICE database [3]).

The number of (qualifying) comparisons recorded in Table 3 must be interpreted differently according to the type of determination that it is linked to: for a wolf it represents the number of false accepts, whereas for a goat it represents the number of false rejects. For example: when using Log-Gabor encoder to generate iris codes of dimension 64x4, the detected *marginal wolf-template* is 334 and it generates 7 cases of impersonation, whereas in the same conditions the *marginal*

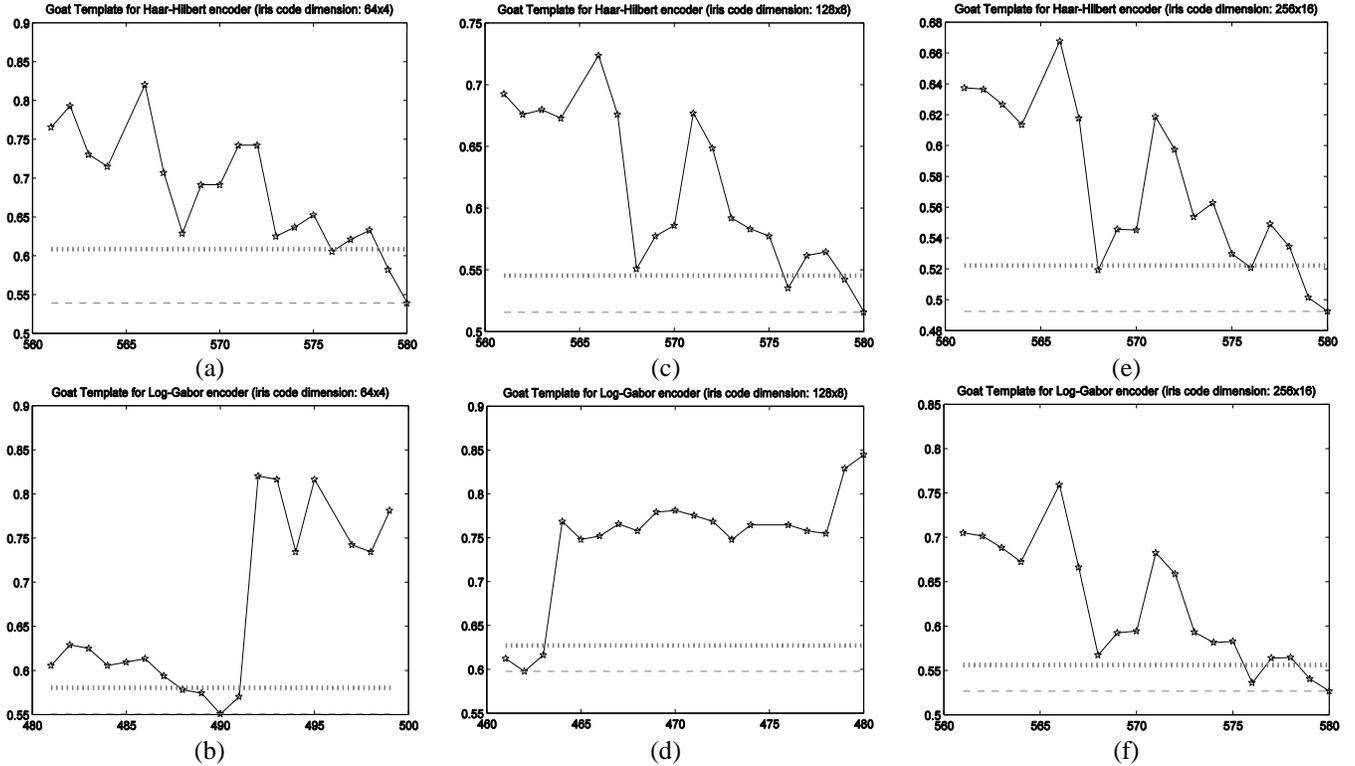

Fig. 2. The *marginal goat-templates* obtained for Haar-Hilbert (64x4 – a, 128x8 – c, 256x16 – e) and Log-Gabor (64x4 – b, 128x8 – d, 256x16 – f) encoders.

*goat-template* is 496 and it generates 4 cases of false reject. What is spectacular in the Table 3 in the first place is that the *marginal goat-template* 496 (Log-Gabor, 64x4) and the *marginal wolf-template* 484 (Log-Gabor, 128x8) point out to the same eye, namely the $25^{th}$ eye, i.e. the left eye of the $13^{th}$ user from the database UBIID, [10]. Section II.C illustrated the fact that trying to qualify users as wolves or goats based on what happens with their template is not quite a simple and evident task. The situation described here reveals an additional degree of difficulty to the same problem, also. Based on the data reported in Table 3, is the left eye of $13^{th}$ user a wolf, a goat or both? This aspect is also a facet of the inconsistency of Biometric Menagerie as a concept.

Fig. 2 illustrates that along with the increasing of the iris code dimension the number of rejections decreases for Log-Gabor encoder and increases for Haar-Hilbert encoder. In each graphic, we drawn the left limit of the safety band (dotted line) and the minimum genuine score (dashed line) obtained for the corresponding *marginal goat template*. Fig. 2.a and Fig. 2.b present the behavior of the *marginal goat-templates* obtained for iris codes of dimension 64x4. The template obtained for Log-Gabor encoder has a bigger number of rejections than the one resulted for Haar-Hilbert encoder. On the contrary, the numbers of rejections for the templates represented in Fig. 2.c and Fig. 2.d are the same for both encoders.

As seen in Fig. 2.e and Fig. 2.f, there are more cases of false reject for the *marginal goat-template* obtained with Haar-Hilbert encoder than for the one obtained with Log-Gabor encoder.

Let us comment another remarkable thing seen in the same Table 3: the *marginal goat-template* obtained for Haar-Hilbert encoder was the same in all three tests. Moreover, it is the *last goat-template* obtained for the same encoder (see Table 4, from below). This situation suggests that the concept of '*goat-template*' could be an objective concept (in certain conditions) unifying the concepts of *first* (*marginal*) and *last goat-templates* by actually depending much on the encoded iris segment and less on the size of the template. The third notable thing visible in Table 3 is that the *marginal wolf-templates* obtained for the six tests were not only different, but also came from different eyes (users). Different iris recognition systems can perceive differently the *marginal wolf-templates*, and consequently, the concept of *marginal wolf-template* is certainly far from being objective.

### E. Detecting the last wolf and goat templates at $t_{EER}$

We recall that the safety levels corresponding to the second series of six exhaustive all-to-all iris recognition tests (further presented here) are those given by running the recognition system at EER threshold $t_{EER}$. Hence, according to the definition 2, each of these tests results in a set containing the *last goat-* and *wolf-templates* corresponding to a given calibration of the biometric system in terms of encoder and iris code size.

Fig. 3 presents the similarity scores obtained by the *last wolf-templates* mentioned in Table 4 and detected in this second series of tests.

As in the previously discussed case of *marginal wolf-templates*, it is visible in Table 4 that the *last*

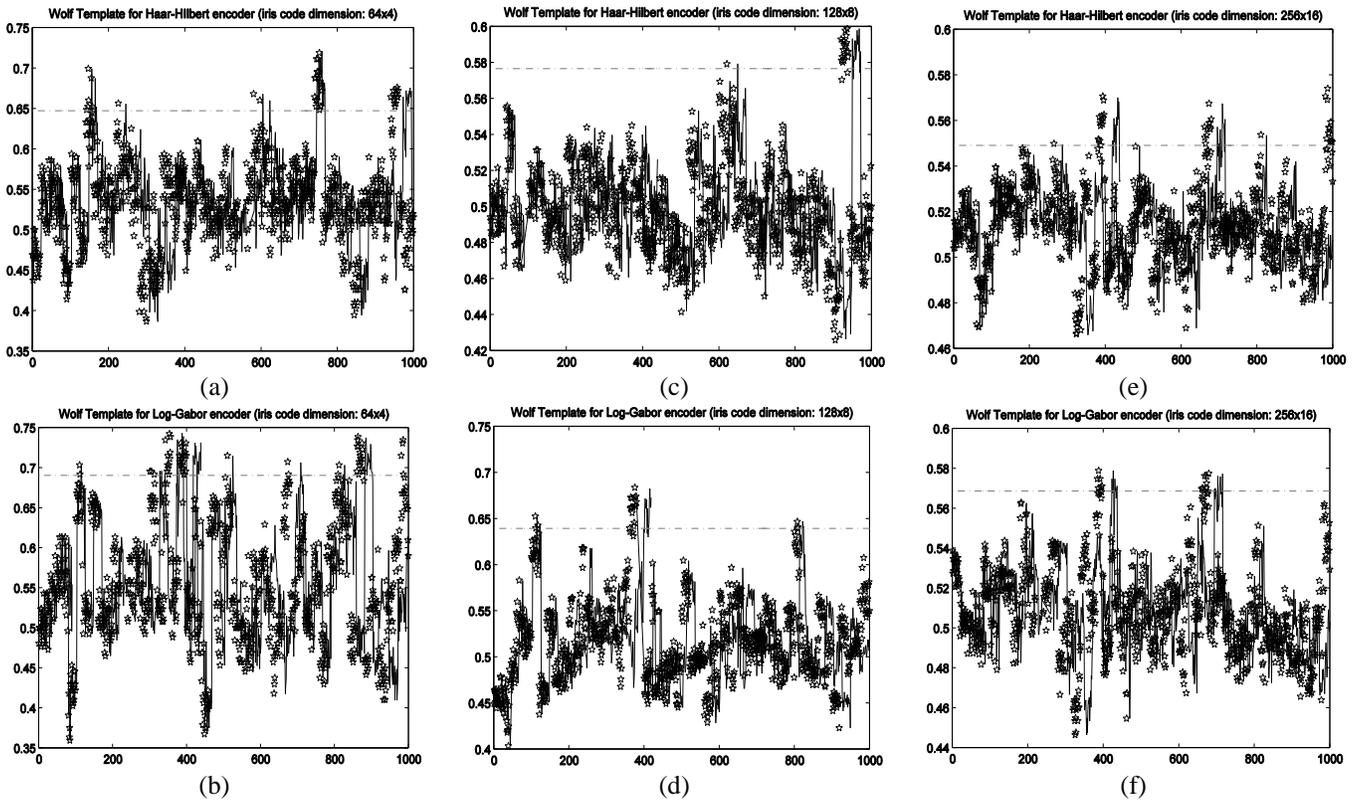

Fig. 3. The similarity scores corresponding to the imposter comparisons generated by *the last wolf-templates* obtained for Haar-Hilbert (64x4 – a, 128x8 – c, 256x16 – e) and Log-Gabor (64x4 – b, 128x8 – d, 256x16 – f) encoders.

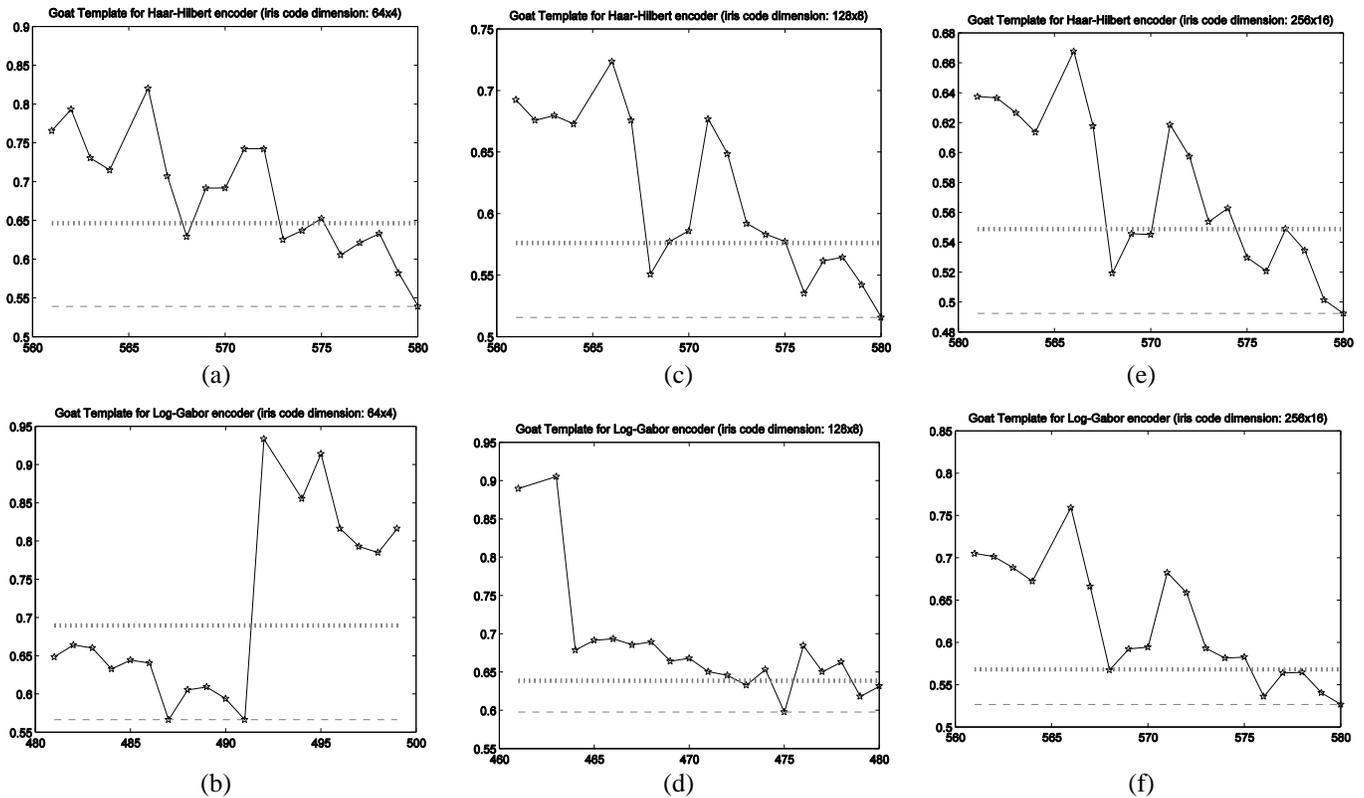

Fig. 4. The similarity scores corresponding to the genuine comparisons generated by the *last goat templates* obtained from the tests that use Haar-Hilbert (iris code dimension: 64x4 – a, 128x8 – c, 256x16 – e) and Log-Gabor (iris code dimension: 64x4 – b, 128x8 – d, 256x16 – e) encoders.

*wolf-templates* obtained for the six tests were not only different, but also came from different eyes (users). Different iris recognition systems can perceive differently the *last wolf-templates*, and consequently, the concept of *last wolf-template* is far from being objective.

TABLE IV
THE *LAST WOLF-/GOAT-TEMPLATES* OBTAINED BY RUNNING THE SYSTEM AT $t_{EER}$

| Iris code dimension<br>Template type | | 64x4<br>Wolf \| Goat | 128x8<br>Wolf \| Goat | 256x16<br>Wolf \| Goat |
|---|---|---|---|---|
| Log-Gabor encoder | Number of comparisons | 63 \| 11 | 22 \| 4 | 14 \| 5 |
| | Template's index | **236 \| 493** | **392 \| 462** | **236 \| 565** |
| Haar-Hilbert encoder | Number of comparisons | 43 \| 8 | 19 \| 6 | 40 \| 9 |
| | Template's index | **549 \| 565** | **88 \| 565** | **236 \| 565** |

TABLE V
THE CUMULATIVE RESULTS OF THE TWO SERIES OF ALL-TO-ALL EXHAUSTIVE IRIS RECOGNITION TESTS (ON UBIID, [10]) EXPRESSED IN TERMS OF *FIRST* AND *LAST GOAT-* AND *WOLF-TEMPLATES*

| Calibration | Goats | | Wolves | |
|---|---|---|---|---|
| | First (Marginal) | Last | First | Last |
| LG, 64x4 | 496 | 493 | 334 | 236 |
| LG, 128x8 | 475 | 462 | 484 | 392 |
| LG, 56x16 | 565 | 565 | 505 | 236 |
| HH, 64x4 | 565 | 565 | 549 | 549 |
| HH, 128x8 | 565 | 565 | 88 | 88 |
| HH, 56x16 | 565 | 565 | 236 | 236 |

TABLE VI
THE CUMULATIVE RESULTS OF THE TWO SERIES OF ALL-TO-ALL EXHAUSTIVE IRIS RECOGNITION TESTS (ON UBIID, [10]) EXPRESSED IN TERMS OF *POSSIBLE FIRST* AND *LAST GOAT-* AND *WOLF-USERS*

| Calibration | Goats | | Wolves | |
|---|---|---|---|---|
| | First (Marginal) | Last | First | Last |
| LG, 64x4 | 25 | 25 | 17 | 12 |
| LG, 128x8 | 24 | 24 | 25 | 20 |
| LG, 56x16 | 28 | 28 | 26 | 12 |
| HH, 64x4 | 29 | 29 | 23 | 23 |
| HH, 128x8 | 29 | 29 | 5 | 5 |
| HH, 56x16 | 29 | 29 | 12 | 12 |

However, there are three different tests pointing out to the template no. 236 (see Table 4) as a *last wolf-template*. Still, this fact alone is not enough for qualifying the concept as being objective. Its extension is strongly dependent on system calibration variables such as the iris code dimension and the texture encoder.

Fig. 4 represents the similarity scores corresponding to the genuine comparisons generated by the *last goat-templates* obtained from the tests that use Haar-Hilbert and Log-Gabor encoders. It illustrates the fact that along with the increasing size of the iris code, the number of false rejects could decrease sometimes.

Table 5 and Table 6 illustrate the cumulative results of the two series of all-to-all exhaustive iris recognition tests (on UBIID, [10]) expressed in terms of *first* and *last goat-* and *wolf-templates* (Table 5), and in terms of *possible first* and *last goat-* and *wolf-users* (Table 6). We said "possible first and last goat- and wolf-users" because, as seen in Section II.C, the process of identifying the wolf users is even fuzzier and more subjective than the process of finding wolf-templates (there is not an unique rule that could qualify users as wolves based on what is happening with their templates). Specifically, the if-then fuzzy rule used here for this purpose is simple as follows:

IF: U *posses a wolf-/goat-template*   THEN: U is a *wolf-/goat-user*.

The data within Table 5 generate the data within Table 6 by applying the above if-then fuzzy rule. The data within both tables allow us to conclude that the goat is the most objective concept of the Fuzzy Biometric Menagerie and Haar-Hilbert encoder is more objective than Log-Gabor encoder.

IV. CONCLUSIONS

This paper shown that, at least in iris recognition, the Biometric Menagerie is a fuzzy and inconsistent concept, regardless if it refers to the users or to their biometric templates. Twelve exhaustive all-to-all iris recognition tests proved this point by counterexample. They also suggest that the goat is the most objective concept of the Fuzzy Biometric Menagerie and that Haar-Hilbert encoder is more objective than Log-Gabor encoder is.

The experimental results presented in this paper shown that the fuzzy-linguistic labels defining the Biometric Menagerie in terms of *wolf-, sheep-, lamb-, goat-users* and those defining the Fuzzy Biometric Menagerie in terms of *first/last wolf-, sheep-, lamb-, goat-templates* or in terms of *possible wolf-, sheep-, lamb-, goat-users*, all of them depend on the calibration of the iris recognition system.

Paradoxically, this paper gave a new perspective on the fuzzy concepts sheep, goats, lambs and wolves, but a very critical one. By illustrating the fact that, different iris recognition systems actually perceive differently the wolf- and goat-templates, the current paper qualifies the concept of Biometric Menagerie as not having one of the most important and most needed attribute of a concept, namely the *universality* with respect to a *genus*.

We wonder if anybody could indicate us a sufficiently large class of iris recognition systems for which the partitioning of the users/templates as a Biometric Menagerie (fuzzy or not) is at least *almost* the same.

Until then, we will remember one of Newton's mottos: *hypotheses non fingo*.

ACKNOWLEDGMENT

The authors would like to thank Professor *Donald Monro* (Dept. of Electronic and Electrical Engineering, University of Bath, UK) for granting the access to the Bath University Iris Image Database.